\documentclass[12pt, final]{l4dc2020}

\usepackage{float}
\usepackage{xcolor}
\usepackage{cleveref}
\usepackage{tikz}
\usepackage{amsmath}
\usepackage{mathtools}
\usepackage{enumerate}
\usetikzlibrary{bayesnet}
\usetikzlibrary{positioning}
\usepackage{hyperref}

\usepackage{multibib}
\newcites{appdx}{Appendix references}

\definecolor{notecolor}{RGB}{0,82,147}
\definecolor{sns1}{HTML}{4C72B0}
\definecolor{sns2}{HTML}{DD8452}
\definecolor{sns3}{HTML}{55A868}
\definecolor{sns4}{HTML}{C44E52}

\newcommand{\dee}{\mathrm{d}}
\newcommand{\GP}{\mathcal{GP}}

\usepackage{pifont}

\title[Planning from Images with Deep Latent Gaussian Process Dynamics]{Planning from Images with Deep Latent Gaussian Process Dynamics}
\usepackage{times}
\usepackage{siunitx}
\newcommand\given[1][]{\:#1\vert\:}

\author{%
 \Name{Nathanael Bosch\nametag{\thanks{Contributed equally \vspace*{-10pt}}
  \addtocounter{footnote}{-1}\addtocounter{Hfootnote}{-1}}} \Email{nathanael.bosch@tum.de}\\
 \addr%
 Max Planck Institute for Intelligent Systems, T\"ubingen, Germany\\
 Technical University of Munich, Germany%
 \AND
 \Name{Jan Achterhold\nametag{\footnotemark}} \Email{jan.achterhold@tuebingen.mpg.de}\\
 \addr%
 Max Planck Institute for Intelligent Systems, T\"ubingen, Germany%
 \AND
 \Name{Laura Leal-Taix\'e} \Email{leal.taixe@tum.de}\\
 \addr%
 Technical University of Munich, Germany%
 \AND
 \Name{J\"org St\"uckler} \Email{joerg.stueckler@tuebingen.mpg.de}\\
 \addr%
 Max Planck Institute for Intelligent Systems, T\"ubingen, Germany%
}

\begin{document}

\maketitle

\vspace{12pt}
\begin{abstract}%
  Planning is a powerful approach to control problems with known environment dynamics.
  In unknown environments the agent needs to learn a model of the system dynamics to make planning applicable.
  This is particularly challenging when the underlying states are only indirectly observable through images.
  We propose to learn a deep latent Gaussian process dynamics (DLGPD) model that learns low-dimensional system dynamics from environment interactions with visual observations.
  The method infers latent state representations from observations using neural networks and models the system dynamics in the learned latent space with Gaussian processes.
  All parts of the model can be trained jointly by optimizing a lower bound on the likelihood of transitions in image space.
  We evaluate the proposed approach on the pendulum swing-up task while using the learned dynamics model for planning in latent space in order to solve the control problem.
  We also demonstrate that our method can quickly adapt a trained agent to changes in the system dynamics from just a few rollouts.
  We compare our approach to a state-of-the-art purely deep learning based method and demonstrate the advantages of combining Gaussian processes with deep learning for data efficiency and transfer learning.
\end{abstract}


\section{Introduction}
Reinforcement learning (RL) has shown success for a number of applications, including
Atari games \citep{DBLP:journals/nature/MnihKSRVBGRFOPB15},
robotic manipulation \citep{DBLP:conf/icra/GuHLL17},
navigation and reasoning tasks \citep{DBLP:conf/icml/OhCSL16},
and machine translation \citep{DBLP:conf/emnlp/GrissomHBMD14}.
Many such results were obtained with model-free deep RL in which the agent directly learns a policy function in the form of a neural network by interacting with the environment.
However, such approaches commonly require a large number of interactions, which often hinders their application to real-world tasks:
Performing actions in real environments, such as driving a vehicle or moving a robot, can be orders of magnitude slower than performing an update of the policy model, and mistakes can carry real-world costs.

Model-based RL is a promising direction to reduce this sample complexity.
In model-based RL, the agent acquires a predictive model of the world and uses the model to make decisions.
This offers several potential benefits over model-free approaches.
First, learning a transition model enables the agent to leverage a richer training signal by using the observed transition instead of just propagating a scalar reward.
Further, the learned dynamics can be independent of the specified task and could therefore potentially be transferred to other tasks in the same environment.
Finally, instead of learning a policy function the agent can use the learned environment for planning to choose its actions.
For environments with only a few state variables, PILCO \citep{Deisenroth11pilco:a} achieves remarkable sample efficiency.
A crucial component is its use of Gaussian processes to model the system dynamics, which allows PILCO to include the uncertainty of the transition model into its policy search.
However, in many problems of interest the underlying state of the world is only indirectly observable through images.
In order to enable fast planning, the agent can learn low-dimensional state representations and model the system dynamics in the learned latent space.
Models of this type have been successfully applied to simple tasks such as balancing cartpoles and controlling 2-link arms \citep{DBLP:conf/nips/WatterSBR15,banijamali17_robus_local_linear_contr_embed}.
However, model-based RL approaches are generally known to lag behind model-free methods in asymptotic performance for problems of this type.
Recently, PlaNet \citep{hafner18_learn_laten_dynam_plann_from_pixel} was able to match top model-free algorithms in complex image-based domains.
PlaNet learns environment dynamics from pixels and chooses actions through online planning in latent space.
Notably, all components in PlaNet are modeled through neural networks.

\subsection{Contributions}
\begin{itemize}\setlength\itemsep{0em}
	\item We combine Gaussian processes (GPs) with neural networks to learn latent dynamics models from visual observations.
	All parts of the proposed deep latent Gaussian process dynamics (DLGPD) model\footnote{ Project page with supplementary material available at \url{https://dlgpd.is.tue.mpg.de/}. \vspace*{-10pt}} can be trained jointly by optimizing a lower bound on the likelihood of transitions in image space.
	\item We integrate the learned system dynamics with learning a reward function and use the models for model-predictive control. In our experiments, the predictions of the learned dynamics model enable the agent to successfully solve an inverted pendulum swing-up task.
	\item We demonstrate that the latent Gaussian process dynamics model allows the agent to quickly adapt to environments with modified system dynamics from only few rollouts. Our approach compares favorably to the purely deep-learning based baseline PlaNet~\citep{hafner18_learn_laten_dynam_plann_from_pixel} in this transfer learning experiment.
\end{itemize}

\subsection{Related Work}
Bayesian nonparametric Gaussian process models \citep{DBLP:books/lib/RasmussenW06} are a popular choice for dynamics models in reinforcement learning (RL) \citep{DBLP:conf/nips/WangFH05,DBLP:conf/nips/RasmussenK03,DBLP:conf/icra/KoKFH07,DBLP:journals/ijon/DeisenrothRP09}.
When the low-dimensional states of the environment are available to the agent, PILCO \citep{Deisenroth11pilco:a} achieves remarkable sample efficiency and is able to solve a swing-up task in a real cart-pole system with only 17.5 seconds of interaction.
Deep PILCO \citep{Gal2016Improving} replace GPs in PILCO with Bayesian neural networks to learn the dynamics model.
The method is not demonstrated to learn an embedding of high dimensional image observations but directly operates on low-dimensional state representations.

Model-free deep RL algorithms have shown good performance in image-based domains \citep{DBLP:journals/nature/MnihKSRVBGRFOPB15,DBLP:journals/corr/LillicrapHPHETS15}, but they commonly require a large number of interactions.
On the other hand, model-based RL can often be more data-efficient.
Many such algorithms learn low-dimensional abstract state representations \citep{DBLP:journals/nn/LesortRGF18} and model the system dynamics in the learned latent space \citep{DBLP:conf/nips/FraccaroKPW17,DBLP:conf/iclr/KarlSBS17}.
Some approaches such as E2C \citep{DBLP:conf/nips/WatterSBR15}, RCE \citep{banijamali17_robus_local_linear_contr_embed} or SOLAR \citep{DBLP:conf/icml/ZhangVSA0L19} learn locally-linear latent transitions and plan for actions based on the linear quadratic regulator (LQR).
In comparison, we learn a non-local, non-linear dynamics model to select actions by planning in latent space.
PlaNet \citep{hafner18_learn_laten_dynam_plann_from_pixel} learns a recurrent encoder and a latent neural transition model to efficiently plan in latent space.
All components of PlaNet are modeled through deep neural networks.
We propose to model state transitions with GPs to reduce the number of model parameters, provide better uncertainty estimates, and generally increase the data-efficiency.

Our formulation provides a data-efficient way to transfer a learned dynamics model to different system dynamics.
Closely related are meta-learning approaches that also learn to transfer between different properties for the same task (e.g.~\citep{alshedivat2018_transferlearning,killian2017_transferlearning,saemundsson2018_metalearning}).
While our models are not specifically trained for transferability, it is an inherent property of our formulation.

\section{Learning Deep Latent Gaussian Process Dynamics for Control}
\label{sec:method}

We propose a novel approach for learning system dynamics from high-dimensional image observations.
We combine the advantages of deep representation learning with Gaussian processes for data-efficient Bayesian modeling of the latent system dynamics.
For control, the agent requires a model of the dynamical system as well as a reward model and an encoder to infer its belief over latent states from observations.
In the following, we formulate our learning framework which learns all these components jointly from image observations and environment interactions.
We also propose our approach for model-predictive control and transfer learning based on our learned models.

\subsection{Deep Gaussian Process State Space Models}

We consider dynamical systems
with latent states
$s \in \mathbb{R}^D$,
actions
$a \in \mathbb{R}^K$,
and observations
$o \in \mathbb{R}^M$.
We call $p(s_{t+1} \given s_t, a_t) = \mathcal{N}( s_{t+1} \given f(s_t, a_t), \Sigma_f )$ the state-transition model with state-transition function $f$ for discrete time steps $t \in \mathbb{N}$.
The observation model $p(o_t \given s_t)$ is associated with the observation function $g$, where $g(s_t)$ is the expected measurement.
In addition, we consider a control task specified by a reward model
$p(r_{t+1} \given s_t, a_t) = \mathcal{N}( r_{t+1} \given h(s_t, a_t), \Sigma_r )$
with reward function $h(s_t, a_t)$ and noise $\Sigma_r$.
\Cref{fig:generative-model} illustrates the generative process.

A possible approach to learn the above models would be to restrict $f$, $g$, and $h$ to families of parametric functions such as deep neural networks.
However, these usually require large training datasets in order to avoid overfitting.
On the other hand, Bayesian nonparametric methods such as GPs often perform well with smaller datasets.
We combine their respective properties and choose different types of methods for the different model components.
%
First, we model the transition function through a GP $f \sim \GP(\mu_f(\cdot), k_f(\cdot, \cdot))$, with mean function $\mu_f: (s_t, a_t) \mapsto s_t$ and radial basis function (RBF) kernel $k_f$.
For state dimensionality $D>1$ we model each output dimension of the transition function with a conditionally independent GP.
Similarly, we chose a GP reward model $h \sim \GP(r_\text{min}, k_h(\cdot, \cdot))$ where $r_\text{min}$ is the minimal reward observed in the collected training data and $k_h$ the RBF kernel.
%
Convolutional neural networks are currently the state-of-the-art method for learning and extracting low-dimensional latent encodings from images. We model the observation function $g$ with a transposed-convolutional network.
For the observation likelihood $p(o_t \given s_t)$ we use an approximate Bernoulli likelihood of $o_t \in [0,1]^{M}$ under the expected measurement $g(s_t)$.
%
Finally, to infer approximate state posteriors from observations we learn a probabilistic encoder of the form $q(s_t \given o_t) \sim \mathcal{N}\left(s_t \given \mu(o_t), \sigma(o_t)^2 \cdot I \right)$ with vector-valued $\mu(\cdot)$ and $\sigma(\cdot)$ parametrized by a convolutional neural network.

To enable state inference from a single observation, we consider observations $o_t$ to consist of two subsequent frames $[i_{t-1}, i_t]$ (see~\cref{fig:observations}).
This allows for approximate inference of positions and velocities, which is sufficient to fully describe the state of the pendulum task considered in our experiments.
Note that for different environments with more complex states or with long-term dependencies it might not be possible to infer a state from two subsequent frames and hence be necessary to choose a more general encoder $q(s_t \given o_{\leq t}, a_{<t})$, e.g. by filtering through a recurrent encoder \citep{hafner18_learn_laten_dynam_plann_from_pixel}.

\subsection{Training Objective}
\label{sec:objective}
We jointly learn all parameters of the model, which include the weights of the neural networks and the hyperparameters of the GP, from interactions with the environment.
Consider transitions $(o_t, a_t, o_{t+1}, r_{t+1})$ collected by interacting with the environment.
To model the covariances of individual data points we group the observed transitions into a joint training dataset
$\mathcal{D} = \left\{ O, A, O', R' \right\}$, with
$O = \{o_1, \dots, o_{T-1}\}$,
$A = \{a_1, \dots, a_{T-1}\}$,
$O' = \{o_2, \dots, o_{T}\}$, and
$R' = \{r_2, \dots, r_{T}\}$.
We further define latent states $S = \{s_1, \dots, s_{T-1}\}$ and $S' = \{s_2, \dots, s_{T}\}$.

With the encoder $q(s_t  \given  o_t)$ we can express the likelihood over latent states as
$q(S \given O) = \prod_{t=1}^{T-1} q(s_t \given o_t)$ and $q(S' \given O') = \prod_{t=2}^{T} q(s_t \given o_t)$.
Similarly, with our observation model we can write
$p(O \given S) = \prod_{t=1}^{T-1} p(o_{t} \given s_{t})$ and
$p(O' \given S') = \prod_{t=2}^{T} p(o_{t} \given s_{t})$.
Finally, since we model the state transitions and the rewards through GPs, the densities $p(S' \given S, A)$ and $p(R' \given S, A)$ are from multivariate Gaussian distributions which we do not factorize over individual transitions.

We marginalize the data likelihood $p(O', R' \given O, A)$ in the following way:
\begin{align}
  p(O', R' \given O, A)
  &= \iint p(O' \given S') \, p(S' \given S, A) \, p(R' \given S, A) \, q(S \given O) \, \dee S \, \dee S'
\label{eq:objective-pre-1}
\\
  &= \mathbb{E}_{q(S' \given O') q(S \given O)} \left[ p(O' \given S') \, p(S'  \given  S, A) \, p(R'  \given  S, A) \, \frac{1}{q(S' \given O')} \right].
  \label{eq:objective-pre-2}
\end{align}
With Jensen's inequality we obtain our training objective as a lower bound on the log-likelihood:
\begin{equation}
  \begin{split}
    \log p(O', R'  \given  O, A)
    &\geq
    \underbrace{
      \mathbb{E}_{q(S' \given O')} \left[\log p(O'  \given  S') \right]
    }_{\text{(I): Reconstruction}}
    +
    \underbrace{
      \mathbb{E}_{q(S' \given O')} \left[- \log q(S' \given O') \right]
    }_{\text{(II): Encoder regularization}}
    \\
    &\quad
    +
    \underbrace{
      \mathbb{E}_{q(S' \given O') q(S \given O)} \left[\log p(S'  \given  S, A) \right]
    }_{\text{(III): State transitions}}
    +
    \underbrace{
      \mathbb{E}_{q(S \given O)} \left[\log p(R'  \given  S, A) \right]
    }_{\text{(IV): Reward}}.
  \end{split}
  \label{eq:objective}
\end{equation}

We refer to \cref{sec:appdx-objective} for a more detailed derivation.
The four terms of the derived lower bound have readily interpretable roles:
The first term (I) describes a (negative) reconstruction loss. Since the decoder parametrizes a Bernoulli distribution over pixel values, it is equivalent to the negative binary cross-entropy loss.
The second term (II) corresponds to the \emph{differential entropy} of the encoder $q(s_t  \given  o_t)$ and can be interpreted as a regularization term on the encoder.
For multivariate Gaussian distribution with diagonal covariance matrix $\Sigma = \operatorname{diag}([\sigma_1^2, \dots, \sigma_D^2])$, the differential entropy is proportional to $\sum_i \log \left( \sigma_i \right)$.
Thus, the term prevents vanishing variances.
Term (III) describes the likelihood of transitions in latent state space in expectation over the encoder.
Since the state transitions are modeled with a GP,
the inner likelihood corresponds to the so-called \emph{marginal log-likelihood} (MLL), which is a common training objective for hyperparameter selection in Gaussian process regressions \citep{DBLP:books/lib/RasmussenW06}.
Similarly, (IV) shows the MLL for the reward-GP, again in expectation over the encoder.
For the loss terms (I), (III), and (IV) we estimate the outer expectations using a single reparametrized sample \citep{kingma13_auto_encod_variat_bayes,rezende14_stoch_backp_approx_infer_deep_gener_model}.
The differential entropy (II) can be computed analytically, since the encoder provides a multivariate Gaussian distribution.

For computational tractability, we maximize the lower bound in eq.~\ref{eq:objective} over batched subsamples of our training dataset, which for Gaussian processes can be theoretically justified by the subset-of-data-approximation \citep{liu18_when_gauss_proces_meets_big_data,DBLP:journals/corr/abs-1901-09541}.
To improve training stability we normalize the encoded states $(S, S')$ batch-wise to zero mean and unit variance before passing them to the transition- and reward GPs.
For testing we compute fixed normalization parameters from the full training dataset.
Before decoding predicted states the normalization is reversed. Furthermore, we limit the signal-to-noise ratio of the RBF kernels of transition and reward GPs by minimizing an additional penalty term (see \cref{sec:snr-penalty}).

\subsection{Posterior Inference with Latent Gaussian Processes}
\label{sec:inferece}
\label{sec:evidence}
\label{sec:conditionals}

To compute the predictive distributions of the transition model and reward model, the GPs need to be conditioned on \emph{evidence}.
With the training data $\mathcal{D} = \left\{ O, A, O', R' \right\}$ and the encoder $q(s \given o)$ we compute latent states $(S, S')$, using reparametrized samples \citep{rezende14_stoch_backp_approx_infer_deep_gener_model, kingma13_auto_encod_variat_bayes} of the predicted distribution as the state representations.
For an arbitrary but known state $s_t^*$ and action $a_t^*$ we then obtain the posteriors
$p \left(s_{t+1}^*  \given  s_t^*, a_t^*, \left( (S,A),S' \right) \right)$ and
$p \left(r_{t+1}^*  \given  s_t^*, a_t^*, \left( (S,A),R' \right) \right)$
through standard posterior GP inference \citep{DBLP:books/lib/RasmussenW06}.

\begin{figure}[t]
  \centering
\subfigure[Image pairs as observations] {
  \resizebox {!} {5em} {
    \label{fig:observations}
    \begin{tikzpicture}
      \begin{scope}
      \node[inner sep=0pt, below=1.3] (img1)
      {\includegraphics[width=.1\textwidth]{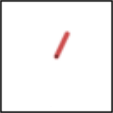}};
      \node [above left,align=left] at (img1.south east){$i_{t-1}$};
      \node[inner sep=0pt, right=1 of img1] (img2)
      {\includegraphics[width=.1\textwidth]{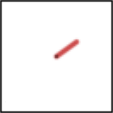}};
      \node [above left,align=left] at (img2.south east){$i_{t}$};
      \node[inner sep=0pt, right=1 of img2] (img3)
      {\includegraphics[width=.1\textwidth]{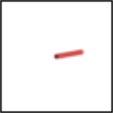}};
      \node [above left,align=left] at (img3.south east){$i_{t+1}$};
      \draw[->] (img1) -- (img2) node[midway,above] {$a_{t-1}$};
      \draw[->] (img2) -- (img3) node[midway,above] {$a_{t}$};
      \node (oi2) [obs, below=1 of img2] {$o_t$};
      \node (oi3) [obs, below=1 of img3] {$o_{t+1}$};
      \edge {img1} {oi2};
      \edge {img2} {oi2};
      \edge {img2} {oi3};
      \edge {img3} {oi3};
      \end{scope}
    \end{tikzpicture} %
  }}\hfil
  \subfigure[Generative model] {
  \resizebox {!} {7.5em} {
  \label{fig:generative-model}
    \begin{tikzpicture}[]
      \begin{scope}
        \node[latent] (s1) {$s_t$};
        \node[obs, below=of s1] (o1) {$o_t$};
        \node[latent, right=of s1] (s2) {$s_{t+1}$};
        \node[obs, above=0.4 of s2] (r2) {$r_{t+1}$};
        \node[obs, above=0.4 of r2] (a2) {$a_{t+1}$};
        \node[obs, left=of a2] (a1) {$a_t$};
        \node[obs, right=of o1] (o2) {$o_{t+1}$};
        \node[latent, right=of s2] (s3) {$s_{t+2}$};
        \node[obs, above=0.4 of s3] (r3) {$r_{t+2}$};
        \node[obs, right=of o2] (o3) {$o_{t+2}$};
        \edge {a1,s1} {s2};
        \edge {a1,s1} {r2};
        \edge {a2,s2} {s3};
        \edge {a2,s2} {r3};
        \edge {s1} {o1};
        \edge {s2} {o2};
        \edge {s3} {o3};
      \end{scope}
    \end{tikzpicture} %
  }}\hfil
  \subfigure[Prediction with DLGPD] {
  \resizebox {!} {9.5em} {
  \label{fig:prediction}
    \begin{tikzpicture}[
      encoder/.style={
        draw, trapezium,
        trapezium stretches=true,
        trapezium left angle=60,
        trapezium right angle=60,
        minimum height=0.8cm,
      },
      decoder/.style={
        draw, trapezium,
        trapezium stretches=true,
        trapezium left angle=60,
        trapezium right angle=60,
        minimum height=0.8cm,
      },
      gp/.style={
        draw, rectangle,
        minimum height=0.8cm,
      },
      ]
      \begin{scope}
      %
      \node (o1) [obs] {$o_t$};
      \node (encoder) [encoder, above=0.49 of o1] {encoder};
      \node (s1) [latent, above=0.46 of encoder] {$s_t$};
      \edge[-] {o1} {encoder};
      \edge {encoder} {s1};
      \node (a1) [obs, above=1.7 of s1] {$a_t$};
      \node (transition) [gp, right=0.8 of s1, minimum width=2.5cm] {transition GP};
      \node (s2) [latent, right=0.8 of transition] {$\hat{s}_{t+1}$};
      \edge[-] {s1} {transition};
      \draw (a1) to[out=-35, in=170] (transition);
      \edge {transition} {s2};
      \node (r2) [latent, above=0.4 of s2] {$\hat{r}_{t+1}$};
      \node (reward) [gp, left=0.8 of r2, minimum width=2.5cm] {reward GP};
      \path (s1) edge[out=30, in=185] (reward);
      \path (a1) edge[out=-30, in=175] (reward);
      \edge {reward} {r2};
      \node (decoder) [decoder, below=0.4 of s2] {decoder};
      \node (o2) [latent, below=0.4 of decoder] {$\hat{o}_{t+1}$};
      \edge[-] {s2} {decoder};
      \edge {decoder} {o2};
      \node (a2) [obs, above=1.7 of s2] {$a_{t+1}$};
      \node (transition2) [right=of s2] {};
      \path (s2) edge[dashed, ->] (transition2);
      \path (a2) edge[dashed, ->, out=-35, in=150] (transition2);
      \node (reward2) [right=of r2] {};
      \path (s2) edge[dashed, ->, out=30, in=200] (reward2);
      \path (a2) edge[dashed, ->, out=-30, in=160] (reward2);
      \end{scope}
    \end{tikzpicture} %
  }}
  \caption{
  	Graphical models depicting relationships between random variables in DLGPD, showing (a) the relation between image frames and observations, (b) the generative model we assume, (c) predicting states and rewards.
  }
\vspace*{-20pt}
\end{figure}
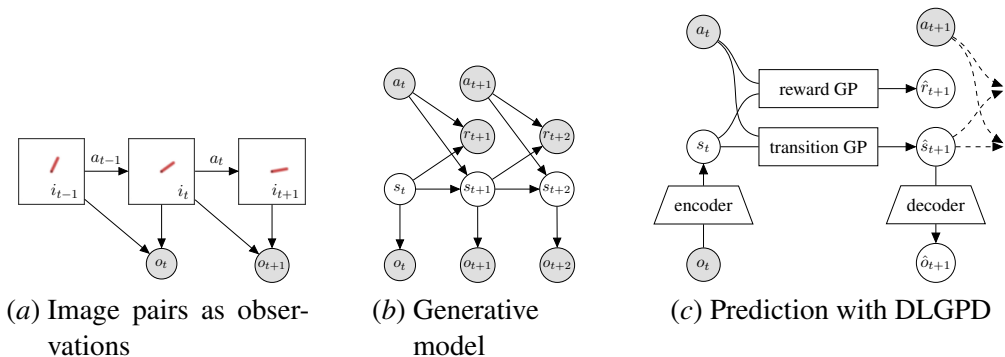

\subsection{Model-Predictive Control}
\label{sec:planning}

We employ our learned system dynamics for model-predictive control (MPC)~\citep{DBLP:journals/automatica/GarciaPM89}.
Key ingredients for MPC are our learned transition and reward models as well as the encoder which infers latent states from image observations.
The observation model is not required for planning. Fig.~\ref{fig:prediction} illustrates prediction with our probabilistic model. We use the cross entropy method (CEM) \citep{Rubinstein96optimizationof,DBLP:journals/anor/BoerKMR05}
to search for the action sequence that maximizes the expected sum of rewards $\mathbb{E}\left[ \sum_{t=1}^T r_t \right]$.
Starting from the mean state encoding of the most recent observation, we compute a state trajectory by forward-propagating the predicted mean.
We then approximate the expected reward by averaging the mean prediction of the reward model for 5 samples of the marginal state distribution for every timestep.

\section{Experiments}
\label{sec:experiments}
We demonstrate our learning-based control approach on the inverted pendulum~(OpenAI Gym Pendulum-v0~\citep{openai_gym}), a classical problem of optimal control and continuous reinforcement learning.
The goal of this task is to swing-up an inverted pendulum from its resting (hanging-down) position.
Due to bounds on the motor torques, a straight upswing of the pendulum is not possible, and the agent has to plan multiple swings to reach and balance the pendulum around the upward equilibrium.

For data collection we excite the system with uniformly sampled random actions $a_t \sim \mathcal{U}([-2, 2])$.
We initialize the system's state with angles $\theta_0 \sim \mathcal{U}([-\pi, \pi])$ and angular velocities $\dot{\theta}_0 \sim \mathcal{U}([-8, 8])$.
We collect 500 rollouts for training and 3 pools of evidence rollouts, containing 200 rollouts each.
Each rollout contains $28$ transitions.

We represent latent states as 3-dimensional real vectors $s_t \in \mathbb{R}^3$.
The observations consist of two subsequent images stacked channel-wise, with image size $64\times64$ pixels and RGB color channels.
We model the probabilistic encoder $q(s_t \given o_t)$ by a convolutional neural network with two output heads for mean and standard  deviation. The decoder $p(o_t \given s_t)$ mapping from the latent space to the observation space is implemented by a transposed-convolutional neural network. For more details on the architecture, see \cref{sec:encoder-decoder}.

For implementation, we use PyTorch \citep{paszke2019pytorch} and GPyTorch \citep{gardner2018gpytorch}.

\subsection{Training}
The lengthscales of the RBF kernel are initialized as $l = \operatorname{softplus}(0) \approx 0.693$.
The outputscales of the transition GPs are initialized to $\alpha^2=1$, the output noise variances to $\sigma^2 = 0.2$.
We pose $\operatorname{Gamma}(1, 5)$ priors on the outputscales and lower bound the outputscales to $10^{-2}$.
For the reward GP, the outputscale $\alpha_\mathrm{reward}^2$ is initialized to the variance of the rewards in the first batch, the output noise variance to $0.2 \cdot \alpha_{\mathrm{reward}}^2$. 
All output noise variances are lower bounded by $\alpha^2 \cdot 10^{-3}$.
The prior mean function for the reward GP is set constant as the minimum of all collected rewards, to predict a minimal reward for unseen regions of the state space.
All parameters of our model are jointly optimized with Adam \citep{kingma2014adam}, with a learning rate of $10^{-3}$ and a batch-size of $1024$.
To encourage the encoder to learn an embedding for the forward modelling task, we stop gradients from the reward model to the encoder.
We train all models for 2000 epochs.
To observe the effect of neural network initialization and training data shuffling, we report control performance on three separately trained models.

We train PlaNet on the same training data like our model ($500$ pendulum rollouts with random initialization and random actions), plus additionally $200$ rollouts which is the maximum number of evidence rollouts we use, which gives $700$ rollouts in total.
Based on the best average performance on 5 validation rollouts, we choose a model after 3.8 million steps of training.
This model serves as the base model for fine-tuning on data from modified environments. For fine-tuning, we evaluate all models every 20k steps for the first 100k steps and every 100k steps for up to 1 million steps, and report results for models where the mean performance is best.

\subsection{Model-Based Control}
\label{sec:mpc}
\begin{figure}[tb]
\centering
\includegraphics[width=1\linewidth]{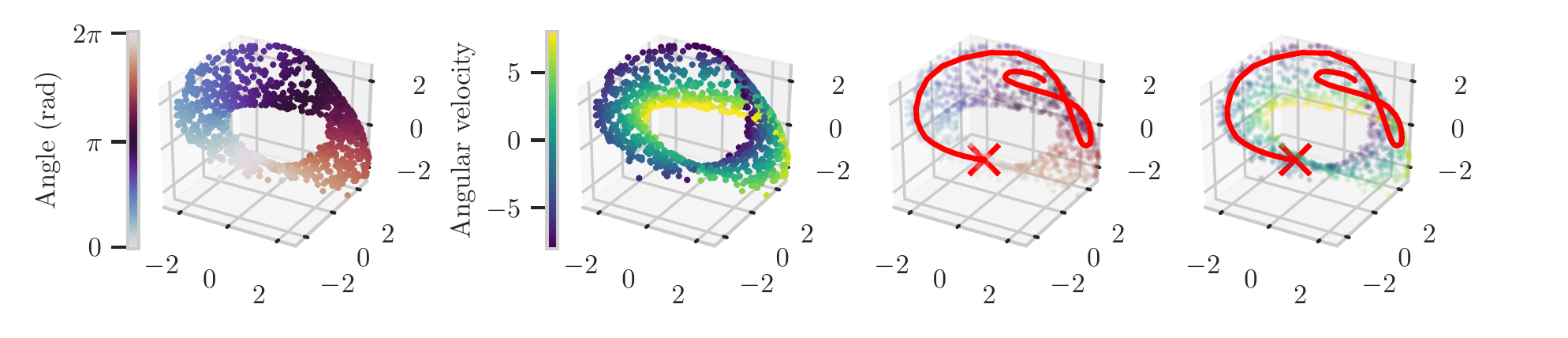}
\vspace*{-30pt}
\caption{Visualizing the 3-dimensional latent space of the learnt embedding. States are colored according to true physical states of the pendulum (angle and angular velocity). On the two rightmost panels, an MPC-planned trajectory for swingup is shown, with \textcolor{red}{\ding{53}} marking the final state.}
\label{fig:pendulum-latent}
\vspace*{-20pt}
\end{figure}

Performance on the swingup task is evaluated on a system randomly initialized with angle $\theta_0 \sim \mathcal{U}([\pi-0.05, \pi+0.05])$ (pole hanging downwards) and angular velocity $\dot{\theta}_0 \sim \mathcal{U}([-0.05, 0.05])$, based on the achieved cumulative reward over $150$ steps.
In order to apply the DLGPD model on prediction and control tasks, the transition and reward GPs have to be conditioned on encoded observations, actions and rewards collected from previous environment interactions. For this we use subsets of rollouts (between 10 and 200 rollouts) from the \textit{evidence} pools.
We use CEM for planning (see \cref{sec:planning}) with a planning horizon of 20 steps.
We report results for 3 control trials on 3 trained models conditioned on a subset of each of the 3 evidence pools (i.e. 27 runs per subset size).
The control performance results in terms of cumulative reward are depicted in fig.~\hyperref[fig:pendulum-performance]{\ref{fig:pendulum-performance}(a)}.
We observe that our approach achieves higher average cumulative reward than PlaNet in this environment already for a small set of evidence rollouts ($\geq 20$).
Fig.~\ref{fig:pendulum-latent} shows a learned latent embedding and a trajectory followed by the CEM planner.

\subsection{Transfer Learning}
\label{sec:transfer}
By modeling the state transitions through a GP, we can learn new state-transition functions of systems with different dynamical properties in a very data-efficient way, as long as the other model components (observation model, reward model, encoder) can be re-used.
In particular, we observed that the agent does not require additional training and that it is sufficient to replace the evidence in the transition GP (see \cref{sec:evidence}) with new data of the modified environment, e.g. collected by following a random policy.
In the following, we investigate the sample efficiency of our model for adapting to environments with changed physical parameters using new random rollouts.
We compare our approach with PlaNet \citep{hafner18_learn_laten_dynam_plann_from_pixel} which needs to be fine-tuned on the same rollouts.

\begin{figure}[tb]
\centering
\includegraphics[width=1\linewidth]{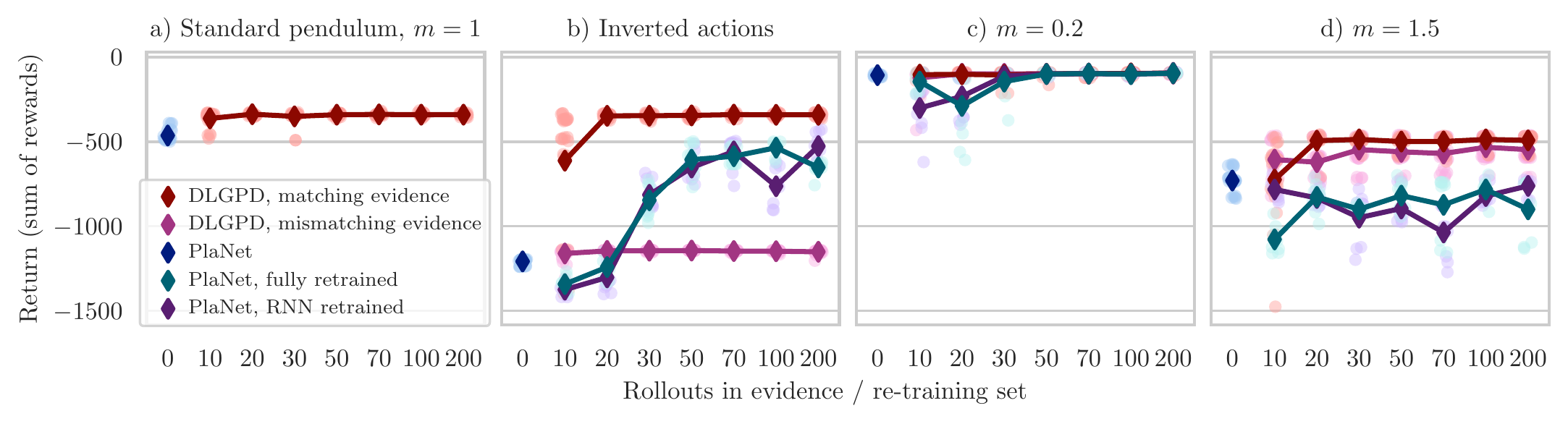}
\vspace*{-30pt}
\caption{Cumulative rewards for PlaNet and our DLGPD model for swing-up of the inverted pendulum in different settings. For the detailed discussion see~\cref{sec:mpc} for (a) and \cref{sec:transfer} for (b-d).}
\label{fig:pendulum-performance}
\vspace*{-20pt}
\end{figure}

To evaluate adaptation capabilities to environments with changed intrinsic properties, we derive three variants of the \textit{Pendulum-v0} environment.
For the \textit{inverted action} environment, we flip the sign of the action before passing it to the original environment.
Second, we make the pole lighter by reducing its mass from $m=1$ to $m=0.2$; we also increase the pole's weight to $m=1.5$.
Results of PlaNet variants and DLGPD conditioned on evidence from the modified and the original environments (matching/mismatching) are shown in fig.~\hyperref[fig:pendulum-performance]{\ref{fig:pendulum-performance}(b-d)}.
For inverted actions, our approach achieves significantly higher cumulative reward even for a small number of rollouts in the evidence.
PlaNet clearly performs less well with the same amount of training data.
With a lower mass than the original pendulum, MPC with both modelling approaches still achieves the swing-up without additional data.
For increased mass, our approach achieves higher cumulative reward than PlaNet with only a few extra rollouts.
In addition to the cumulative reward, we also evaluated our experiments with respect to the ratio of successful rollouts (see \cref{sec:pendulum-success-rate}).


\section{Conclusions}

We propose DLGPD, a dynamics model learning approach which combines deep neural networks for representation learning from images with Gaussian processes for modelling dynamics and rewards in the latent state representation.
We jointly train all model parameters from example rollouts in the environment and demonstrate model-predictive control on the inverted pendulum swing-up task.
Our latent GP transition model allows for data-efficient transfer to tasks with modified pendulum dynamics without additional training, by conditioning the transition GP on rollouts from the modified environment.
In comparison to a state-of-the-art purely deep learning based approach (PlaNet) our method demonstrates superior performance in data-efficiency for transfer learning.

Scaling and evaluating our approach on more complex tasks, environments and eventually robotic systems is an interesting topic for future research.
To this end, we see large potential in collecting additional training data intermittently during model training, as in PILCO~\citep{Deisenroth11pilco:a} and PlaNet~\citep{hafner18_learn_laten_dynam_plann_from_pixel}.
Further potential is in the use of more complex or recurrent encoder architectures. 
Moreover, a probabilistic treatment of the forward prediction, e.g. through moment-matching or Monte-Carlo sampling, might further improve data-efficiency.

\acks{This work has been supported through the Max Planck Society and Cyber Valley. The authors thank the International Max Planck Research School for Intelligent Systems (IMPRS-IS) for supporting Jan Achterhold.
}

\bibliography{references}

\newpage
\appendix
\numberwithin{equation}{section}
\renewcommand{\theequation}{I\thesection.\arabic{equation}}

\section{Lower bound derivation}
\label{sec:appdx-objective}
In this section we derive a lower bound on the transition log-likelihood $\log p(O', R' \given O, A)$ in more detail compared to \cref{sec:objective}, where the corresponding symbols and notation is introduced.
Equations \ref{eq:objective-pre-1}-\ref{eq:objective} are revisited here and referred to by their original equation number.
We first marginalize over latent states $S$, $S'$
\begin{align}
  p(O', R' \given O, A)  &= \iint p(O', R', S, S' \given O, A) \, \dee S \, \dee S' \\
\intertext{and factorize}
  p(O', R' \given O, A)  &= \iint p(O', S' \given S, O, A) p(R' \given S, O, A) p(S \given O, A) \, \dee S \, \dee S', \\
\intertext{which can further be simplified by applying conditional independence assumptions}
  &= \iint p(O' \given S') p(S' \given S, A) p(R' \given S, A) p(S \given O) \, \dee S \, \dee S'.
 \tag{\ref{eq:objective-pre-1}} \\
\intertext{By introducing a variational approximation $q(S' \given O')$ and renaming $p(S \given O) \rightarrow q(S \given O)$ to highlight its role as an encoder, we finally arrive at the expectation}
  p(O', R' \given O, A)
  &= \mathbb{E}_{q(S' \given O') q(S \given O)} \left[ p(O' \given S') \, p(S'  \given  S, A) \, p(R'  \given  S, A) \, \frac{1}{q(S' \given O')} \right].
  \tag{\ref{eq:objective-pre-2}}
\intertext{With Jensen's inequality we obtain our training objective as a lower bound on the log-likelihood:}
  \begin{split}
    \log p(O', R'  \given  O, A)
    &\geq
    \underbrace{
      \mathbb{E}_{q(S' \given O')} \left[\log p(O'  \given  S') \right]
    }_{\text{(I): Reconstruction}}
    +
    \underbrace{
      \mathbb{E}_{q(S' \given O')} \left[- \log q(S' \given O') \right]
    }_{\text{(II): Encoder regularization}}
    \\
    &\quad
    +
    \underbrace{
      \mathbb{E}_{q(S' \given O') q(S \given O)} \left[\log p(S'  \given  S, A) \right]
    }_{\text{(III): State transitions}}
    +
    \underbrace{
      \mathbb{E}_{q(S \given O)} \left[\log p(R'  \given  S, A) \right]
    }_{\text{(IV): Reward}}.
  \end{split}
  \tag{\ref{eq:objective}}
\end{align}

\section{Signal-to-noise ratio (SNR) regularization}
\label{sec:snr-penalty}
As covariance functions (between targets) for the transition GPs and reward GP we choose radial basis function (RBF) kernels
\begin{equation}
k(x_i, x_j) = \alpha^2 \exp \left(-\frac{1}{2}(x_i - x_j)^\mathrm{T} \Lambda^{-1} (x_i - x_j) \right) + \delta_{ij} \sigma^2
\end{equation} with outputscale $\alpha > 0$, additive noise covariance $\sigma^2$, characteristic length-scales $\Lambda = \mathrm{diag}([l_1^2, ..., l_D^2])$ and indicator function $\delta_{ij}$ ($\delta_{ij} = 1 \text{ if } i=j, \delta_{ij}=0$ otherwise). To improve numerical stability, we regularize the ratio of outputscale and noise covariance of the RBF kernels of the transition GPs (denoted by $\mathcal{K}_\mathrm{trans}$) and reward GP (denoted by $K_\mathrm{reward}$) by minimizing a signal-to-noise penalty
\begin{equation}
\mathcal{L}_\mathrm{SNR} = \sum_{k \: \in \:  \mathcal{K_\mathrm{trans}} \cup K_\mathrm{reward} } \ \left[ \frac{\log(\mathrm{SNR}_k)}{\log(\tau)} \right] ^ p
\end{equation} with $\mathrm{SNR}_k = \alpha_k / \sigma_k, \tau = 10, p = 8$. A similar regularization can be found in the PILCO implementation \citepappdx{deisenroth_pilco_code} (with $\tau = 1000, p = 30$). The final loss we minimize is thus
\begin{equation}
\mathcal{L} = -\mathcal{L}_\mathrm{lower-bound} + \mathcal{L}_\mathrm{SNR}
\end{equation} with
\begin{equation}
 \begin{split}
    \mathcal{L}_\mathrm{lower-bound}
    &=
      \mathbb{E}_{q(S' \given O')} \left[\log p(O'  \given  S') \right]
    +
      \mathbb{E}_{q(S' \given O')} \left[- \log q(S' \given O') \right]
    \\
    &\quad
    +
      \mathbb{E}_{q(S' \given O') q(S \given O)} \left[\log p(S'  \given  S, A) \right]
    +
      \mathbb{E}_{q(S \given O)} \left[\log p(R'  \given  S, A) \right]
  \end{split}
\end{equation}
(see \cref{eq:objective}).

\section{Control rollout success rate}
\label{sec:pendulum-success-rate}

\begin{figure}[H]
\centering
\includegraphics[width=1\linewidth]{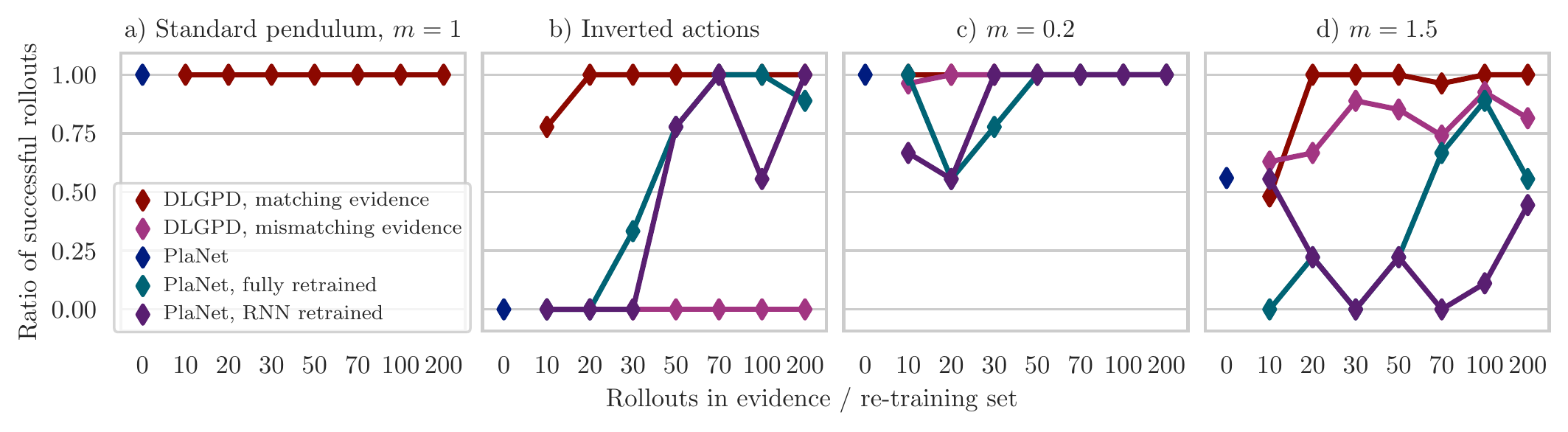}
\vspace*{-30pt}
\caption{In addition to the cumulative rewards collected for pendulum swingup rollouts with our DLGPD model and PlaNet (see \cref{sec:transfer}) we also evaluated the ratio of successful control rollouts for several environment modifications and evidence (DLGPD) / re-training (PlaNet) subset sizes.
As there is no common definition of "success" for the \textit{Pendulum-v0} environment of OpenAI Gym, we defined a rollout to be successful if the last 25 planning steps of 150 total steps exceed a reward of $-1$ ($0$ is the maximum achievable reward for this environment).
The experimental setting is otherwise analogous to the evaluation in \cref{sec:experiments}.
We observe that DLGPD matches or outperforms the success rate of PlaNet for evidence sizes $\geq$~20 in all settings and is able to adapt to changes in the environment (\cref{fig:pendulum-success-rate} (b-d)) with only a few extra rollouts.
}
\label{fig:pendulum-success-rate}
\vspace*{-20pt}
\end{figure}

\clearpage
\section{Encoder and decoder architecture}
\label{sec:encoder-decoder}
For mapping from observation space to latent space and vice versa, we use (transposed) convolutional neural networks with ReLU non-linearities. The network architecture is similar to the encoder and decoder used in \citep{hafner18_learn_laten_dynam_plann_from_pixel}. The encoder parametrizes a normal distribution by its mean $\mu$ and standard deviation $\sigma$.
\begin{table}[h]
\caption{Encoder architecture}
\center
\begin{tabular}{| l |}
\hline
Input: 2 channel-wise concatenated RGB images (6x64x64) \\
\hline
Conv2D (32 filters, kernel size 4x4, stride 2) + ReLU \\
\hline
Conv2D (64 filters, kernel size 4x4, stride 2) + ReLU \\
\hline
Conv2D (128 filters, kernel size 4x4, stride 2) + ReLU \\
\hline
Conv2D (256 filters, kernel size 4x4, stride 2) + ReLU \\
\hline
$\mu$: Linear (1024 $\rightarrow$ 3), $\sigma$: Softplus(Linear(1024 $\rightarrow$ 3) + 0.55) + 0.01 \\
\hline
\end{tabular}
\end{table}
\begin{table}[h]
\caption{Decoder architecture}
\center
\begin{tabular}{| l |}
\hline
Input: 3-dimensional latent variable \\
\hline
Linear(3 $\rightarrow$ 1024) + ReLU \\
\hline
ConvTranspose2D (128 output channels, kernel size 5x5, stride 2) + ReLU \\
\hline
ConvTranspose2D (64 output channels, kernel size 5x5, stride 2) + ReLU \\
\hline
ConvTranspose2D (32 output channels, kernel size 6x6, stride 2) + ReLU \\
\hline
ConvTranspose2D (6 output channels, kernel size 6x6, stride 2) + Sigmoid \\
\hline
Output: 2 channel-wise concatenated RGB images (6x64x64) \\
\hline
\end{tabular}
\end{table}

\bibliographystyleappdx{plainnat} 
\bibliographyappdx{references}

\end{document}